\documentclass[10pt]{article}
\usepackage[utf8]{inputenc}
\usepackage[
  left=1.5in,
  right=1.5in,
  top=1in,
  bottom=1in
]{geometry}
\usepackage{url}
\usepackage[T1]{fontenc}
\usepackage{newtxtext}
\usepackage{amsfonts}
\usepackage{booktabs}
\usepackage{tabularx}
\usepackage{xurl}
\usepackage{graphicx}
\usepackage{amsmath}
\usepackage{newtxmath}
\usepackage{float}
\usepackage{indentfirst}
\usepackage{caption}
\usepackage{xcolor}
\usepackage{titlesec}
\usepackage{fancyhdr}
\fancypagestyle{firstpage}{
  \fancyhf{}
  \fancyfoot[C]{\thepage}
  
}

\titleformat{\section}
  {\fontsize{12}{14}\selectfont\bfseries}
  {\thesection}
  {0.6em}
  {}

\titleformat{\subsection}
  {\fontsize{10.5}{12.5}\selectfont\bfseries}
  {\thesubsection}
  {0.6em}
  {}

\titlespacing*{\section}
  {0pt}{2.5ex plus 0.5ex minus 0.2ex}{1.2ex}

\titlespacing*{\subsection}
  {0pt}{2ex plus 0.4ex minus 0.2ex}{0.8ex}

\usepackage{hyperref}
\hypersetup{
  pdftitle={Metacognitive Steering: Learning the Structure of Scientific Judgment},
  pdfauthor={Autopoiesis Sciences, Inc.}
}

\begin{document}
\thispagestyle{firstpage}
\vspace*{0.15in}
\begin{center}
  \includegraphics[width=3.5cm]{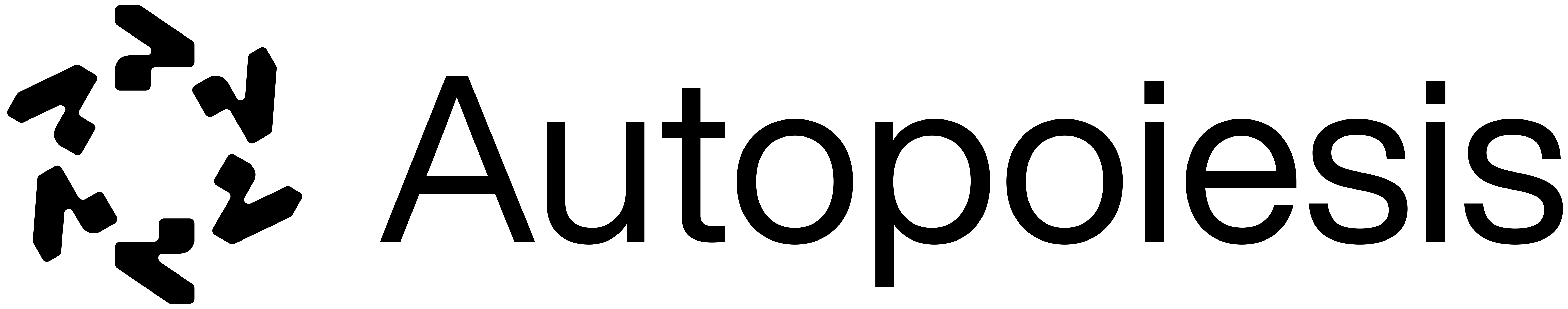}\\[1em]

  \rule{\textwidth}{4pt}\\[1.8em]

  \begin{minipage}{\textwidth}
    \centering
    {\fontsize{16}{17.5}\selectfont\bfseries
    Metacognitive Steering: Learning the Structure of Scientific Judgment
    \par}
  \end{minipage}

  \vspace{1.8em}
  \rule{\textwidth}{1.25pt}

  \vspace{1.5em}

  {\fontsize{9.5}{11.5}\selectfont\bfseries
  Vincent Karpf, Joseph Reth, Eike Gerhardt, Audrey Wang,\\
  Anna Butz, Jiehao Xing, Jialing Song, Larry Callahan
  \par}

  \vspace{0.4em}
  {\fontsize{9.5}{11}\selectfont Autopoiesis Sciences\par}
  {\fontsize{8.5}{10}\selectfont\ttfamily
  research@autopoiesis.science\par}
\end{center}

\vspace{2.2em}

\begin{center}
  \begin{minipage}{0.84\textwidth}
    \begin{center}
      {\fontsize{11}{13}\selectfont\bfseries Abstract}
    \end{center}

    \vspace{0.6em}
    \small
Long-horizon scientific discovery requires agents to alternate between
exploration, disciplined execution, and critical reassessment as evidence
changes. Current language models are trained primarily on the products of
science and optimized using outcome-level signals, providing limited supervision
for these process-level shifts in scientific judgment. We investigate whether
such judgment can be recovered from scientist interaction traces and used to
control the internal computation of a frozen frontier model. Using
contrastive interventions collected during real scientific research, we
identify a coordinated, low-dimensional control structure within Kimi~2.6, a
trillion-parameter mixture-of-experts model. Residual analysis,
attention-weight subspace alignment, and cross-layer singular value decomposition
converge on a mid-depth control surface spanning key layers. We
introduce Metacognitive Steering, an inference-time controller that reads the
model's cognitive regime and dynamically composes layer-specific interventions
for exploration, procedural convergence, or critical reassessment without
modifying model parameters. Behavioral analyses show that this control produces
more sustained exploration, explicit pruning, and evidence-responsive
synthesis. We operationalize the method in Columbus-1, an autonomous research
system that identified eight independently reproduced, attacker-reachable
vulnerabilities in BlueZ and directed the design, simulation, and fabrication
of a ten-foot rocket intended to land propulsively using non-throttleable solid
motors. Together, these results show that process-level scientific judgment can
provide supervision for interpretable, dynamic control over a model's reasoning
strategy.
  \end{minipage}
\end{center}

\vspace{1.5em}

\section{Introduction}

The next leap in AI for science may come not from a bigger model, but from a new kind of data: the reasoning scientists discard. Recent research identifies a gap between scientific workflow execution and evidence-responsive reasoning. Across more than 25,000 scientific-agent runs, agents ignored available evidence in 68\% of traces, revised beliefs following refutation in only 26\%, and rarely assembled convergent evidence across multiple tests~\cite{riosgarcia2026scientists}. These findings suggest that current systems can execute scientific workflows without consistently following the epistemic processes that make science self-correcting.
We argue that a central bottleneck is not only model capability, but a supervision gap: the absence of training data that capture reasoning under scientific uncertainty. Process-level records from working scientists, including intermediate reasoning, failed experiments, discarded alternatives, and belief revisions, are largely absent from the published record~\cite{kumbhar2026cogtrl}. Such records may provide training signals that parameter scaling alone does not. This limitation is particularly visible in the prevailing training paradigm: reinforcement learning is most effective when success can be cheaply and reliably verified. That bet has been especially successful in mathematics and coding, where a verifiable reward can be computed at scale~\cite{deepseekai2025r1}. A proof can be checked~\cite{hubert2026nature}. Code can be executed. A game has a finite board and an objective score. When verification is inexpensive, models can generate many candidate solutions and receive direct feedback from success or failure. Scientific discovery is different.

The ease of training AI on a task scales with how verifiable the task is~\cite{wei2025verifiers}, and scientific hypotheses sit at the wrong end of this asymmetry: easy to propose, expensive to confirm. Experiments are costly, and the significance of what they return may not be clear for years. A failed experiment may falsify a hypothesis or merely expose an unmodeled variable, and nothing in the result itself says which. Under a working paradigm such results are often treated as puzzles still to be solved, so deciding which interpretation warrants further investigation is a problem of judgment under uncertainty rather than a score~\cite{kuhn1962structure}. Language models struggle with that same demand: they often fail to revise or retain beliefs when new evidence arrives~\cite{wilie2024belief,riosgarcia2026scientists}, and they lack the metacognitive control that reasoning under uncertainty requires~\cite{griot2025metacognition}. Scientists do not reason in one continuous mode. They explore when evidence is weak, proceed methodically when a working framework holds, and challenge assumptions when results cease to make sense~\cite{kuhn1962structure}. They decide which weak signal is worth following before certainty exists. Such shifts in reasoning mode may be central to scientific discovery, yet the conditions under which they arise remain poorly recorded. Current text-trained language models largely inherit the products of science without inheriting the process that produced them~\cite{kumbhar2026cogtrl}: a paper records the path ultimately taken, not the competing hypotheses, the reversals, or the moments of doubt that produced it. Consequently, a model trained primarily on published text learns how science is written after uncertainty has been resolved, rather than how science is done while the answer remains unknown~\cite{felin2024theory}.

If process signal can be captured, it can also be used to control a model directly. Activation Addition and Contrastive Activation Addition showed that the behavior of a frozen model can be shifted by adding a direction to its residual stream~\cite{turner2023activation,rimsky2024caa}. Representation Engineering developed the broader program of reading and writing these internal directions~\cite{zou2023representation}. Later work extended residual-stream control to context-gated interventions, to reasoning behaviors such as uncertainty and backtracking, and to sparse neural signals predictive of reasoning outcomes~\cite{lee2025cast,venhoff2025reasoning,dong2026neurons}. Subsequent methods make the intervention itself dynamic or context-conditioned, composing or gating directions as the input changes~\cite{scalena2024dynamic,wang2025sadi}. Those controllers still typically target one behavior with one family of directions, and nearly all have been demonstrated at 7B to 70B parameters. Scientific reasoning instead asks for dynamic control of cognitive stance, which requires both a set of directions to choose from and a signal for when each one applies. We derive the directions from a proprietary dataset of process-level scientist interventions recorded during real long-horizon research, each paired with a counterfactual that holds the scientific context fixed and changes only the cognitive stance. These pairs make a new form of control possible. We introduce Metacognitive Steering, a closed-loop controller that infers the model's current reasoning state from its own residual stream and applies the direction that state calls for as work moves between exploration, convergence, and scrutiny. On a trillion-parameter open model, we localize a mid-depth control surface where this variation is written and read, and show that steering it changes how the model reasons.

\section{Dataset Preparation}

The record of scientific failure is rarely preserved in any structured, machine-readable form. This is not a matter of secrecy. It is simply not the kind of thing a scientist has reason to formalize, so it stays scattered across lab notebooks, half-remembered conversations, and abandoned analysis notebooks. This is a data problem.
To capture and learn from this otherwise inaccessible scientific knowledge, we built an AI Co-Scientist with several models purpose-built for specific phases in the scientific method. The system accelerates their work, searching literature, generating hypotheses, and analyzing results, while the interaction captures something that published literature removes: scientific judgment in motion.

Verified researchers are using the system on novel, long-horizon research problems. Their traces contain experimental data and both stated and revealed preferences. A scientist may explicitly ask the system to challenge an assumption. Just as importantly, they may ignore one result, revisit another, redirect the investigation, narrow a hypothesis, reject a plausible explanation, or continue working when the immediate evidence is weak. Over long horizons, those decisions reveal what scientists attend to, what they distrust, and what they believe is worth pursuing. This forward-looking, theory-guided character of scientific reasoning, in which beliefs shape which data is sought rather than merely being updated by it, is precisely what distinguishes human cognition from the backward-looking predictive logic of current AI systems~\cite{felin2024theory}.

From these sessions, we constructed a proprietary judgment dataset containing process-level scientist interventions with multi-turn research trajectories and selected graph memory as context. We filtered the dataset using an LLM-based classification pipeline, an approach increasingly validated as a substitute for human annotation on complex labeling tasks~\cite{ziems2024llms}. The pipeline assigns each intervention to one of six categories: hypothesis generation or revision, tension identification between evidence or hypotheses, strategic pivot, assumption surfacing, result interpretation, or experimental design. The classifier prompt was refined through iterative optimization cycles against a human-labeled subset of real scientist interventions: evaluation failures were identified, targeted fixes applied, and the revised prompt rescored against the labeled set, following the prompt optimization with human feedback paradigm~\cite{lin2024pohf}.

We then generated synthetic negative interventions that hold the scientific context constant while changing the cognitive stance of the scientist. This contrastive-pair construction follows the extraction methodology of activation steering, where behavioral directions are derived from paired inputs that differ in stance while controlling for content~\cite{venhoff2025reasoning,dong2026neurons}. An LLM verifier scored expected content overlap and we discarded low-overlap pairs. On the retained pairs, a true positive--negative pair was 3.6 times more similar in word-and-bigram TF-IDF cosine than a random re-pairing of that scientist's own negatives onto their positives (one-sided permutation $p = 0.001$; bootstrap 95\% CI 3.3--3.8). In the final dataset, the intended difference is not primarily what subject to discuss, but how to think about it: explore rather than converge, scrutinize rather than accept, operationalize rather than speculate.

This distinction is essential. A collection of expert answers can teach a model what experts say. Interaction data can begin to teach it when experts change stance. Prior work has shown that structured training data extracted from authentic expert scientific discussion improves model performance on scientific reasoning~\cite{yin2025scientific}, and that intent and action structure in expert dialogue is learnable and evaluable as supervision signal~\cite{chen2025act}.

The value of this dataset is its structure:

\begin{enumerate}
  \item It is generated organically during real scientific work rather than through decontextualized annotation.
  \item It preserves unresolved uncertainty rather than only successful conclusions.
  \item It captures negative knowledge: rejected ideas, failed searches, weak evidence, hidden constraints, and reasons not to proceed.
  \item It records decisions across long horizons, where the consequences of early judgment become visible later.
  \item It comes from qualified scientists working on problems whose answers are not already contained in the training corpus.
  \item It contains interaction metadata like dwell time, clicked paper links, whether they shared the response, notes taken, highlights, branch decisions, and response revisions
\end{enumerate}

This is the type of data required to build systems that do more than retrieve and summarize scientific knowledge. It is process-level supervision for an underdeveloped capability in autonomous systems: forming models of an uncertain problem and deciding what to investigate next.

\section{Interpretability Research and Proposed Method}
\subsection{Locating the Control Surface of Scientific Reasoning}

We first set out to determine where cognitive stance is established in a frontier model and how it propagates through the network. All experiments use Kimi~2.6~\cite{moonshot2026kimi}, a DeepSeek-V3-style
transformer~\cite{deepseekai2024v3} with 61 layers, residual stream width
$d = 7168$, and 64 multi-head latent attention (MLA)
heads~\cite{moonshot2026kimi}. The model is a trillion-parameter
mixture-of-experts architecture; because our analyses operate on residual
streams and attention projections, we treat the MoE structure as architectural
context and make no claims about expert-level routing in this section.

Our aim is to measure variation in \emph{cognitive stance}: how the model is
thinking about a problem, holding the problem itself fixed. Measuring this
directly is not possible from ordinary generations, because differences in
output conflate topic, style, and cognition. We therefore adopt the contrastive
extraction methodology of activation
steering~\cite{turner2023activation,rimsky2024caa} and generate paired rollouts
in which the scientific context remains fixed and only the cognitive
intervention changes. For each pair and each layer $\ell$, we compute the
difference of residual activations,
\begin{equation}
\Delta h^{(\ell)} \;=\; h_{\mathrm{pos}}^{(\ell)} \;-\;
h_{\mathrm{neg}}^{(\ell)},
\end{equation}
where $h^{(\ell)}$ denotes the residual activation at the final generated
token of the rollout, extracted identically across all layers and analyses in
this paper. These difference vectors isolate changes associated with how the model reasons while controlling for the scientific problem.

\begin{figure}[H]
  \centering
  \begin{minipage}[t]{0.485\textwidth}
    \centering
    \includegraphics[width=\linewidth]{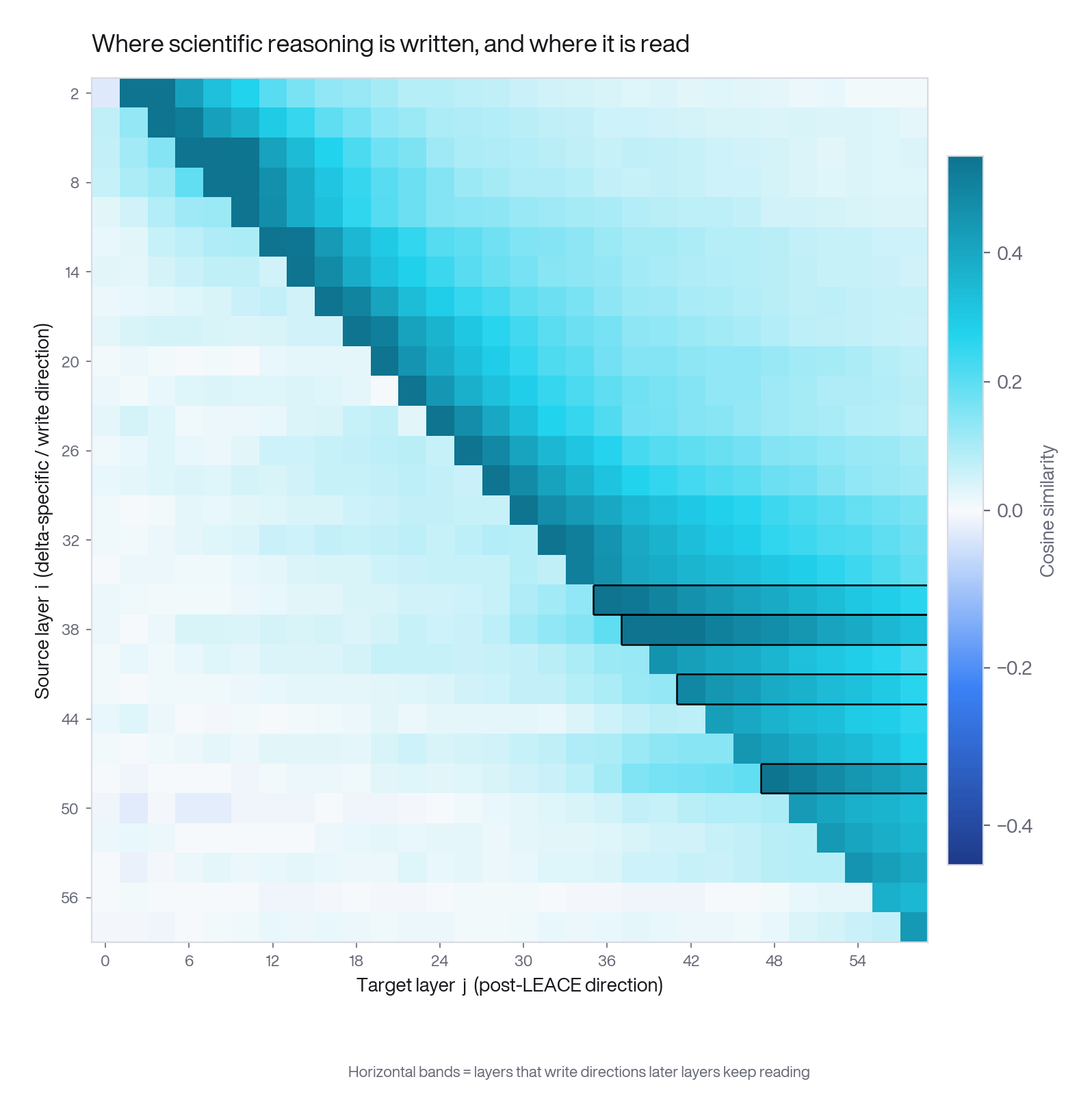}
    \caption{Post-erasure residual-stream alignment. Each cell reports
    $A_{ij}$; outlined horizontal bands identify source layers with strong
    alignment to later-layer directions.}
    \label{fig:residual}
  \end{minipage}
  \hfill
  \begin{minipage}[t]{0.485\textwidth}
    \centering
    \includegraphics[width=\linewidth]{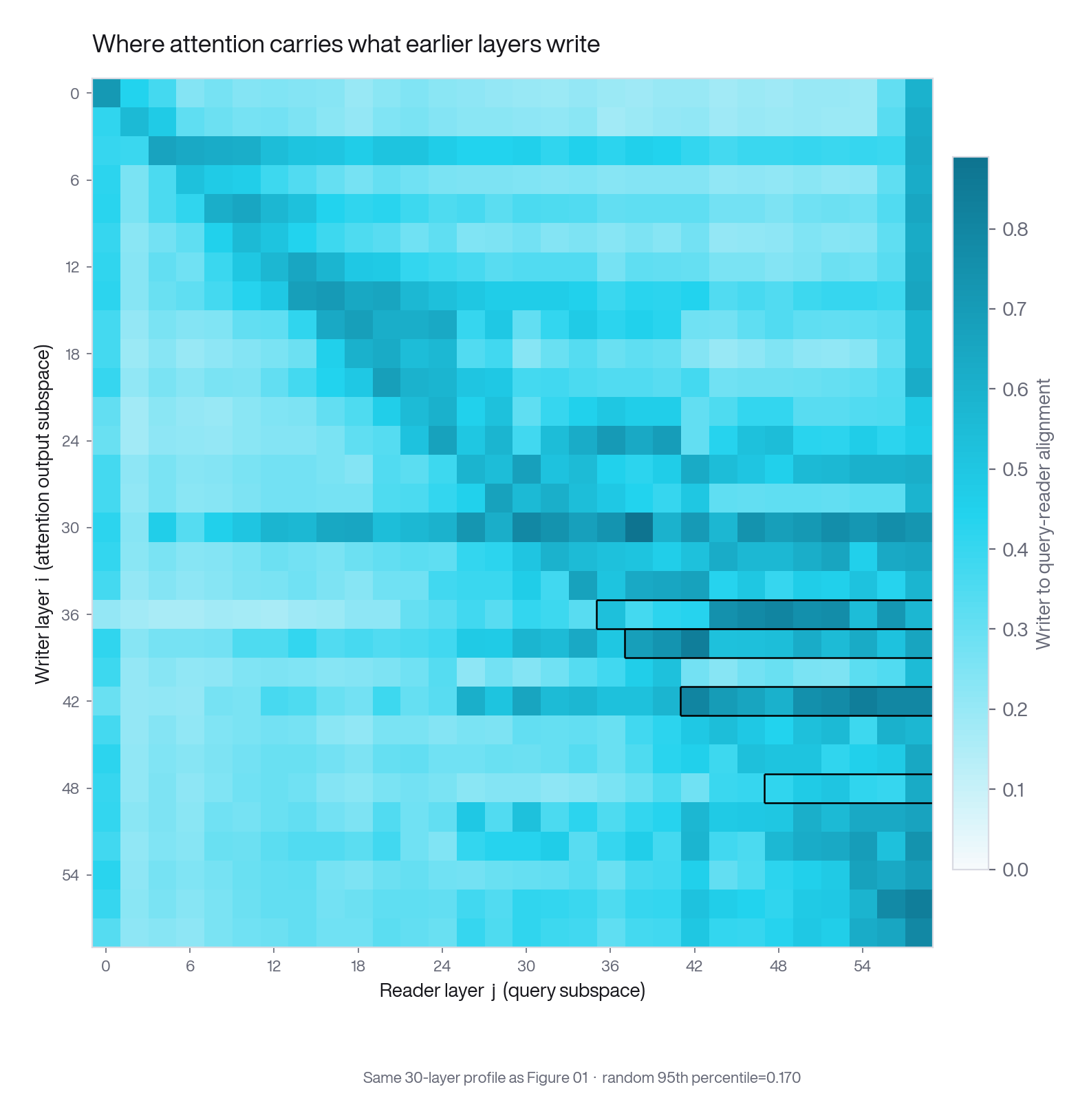}
    \caption{Frozen-weight attention write-to-query alignment. Each cell
    reports $B_{ij}$ for $r=50$; the 95th percentile of the Haar-random
    subspace null is $0.170$.}
    \label{fig:attention}
  \end{minipage}
\end{figure}

A first finding of this analysis is also a hypothesis about what it contains:
contrastive interventions share a dominant component related to general
instruction following. All pairs differ along a common ``an intervention
occurred'' axis, because every rollout was prompted to change something; this
shared component is not scientific cognition but an artifact of the
manipulation itself. Without correcting for it, an alignment analysis would
rediscover that generic signal at every layer. We therefore apply a rank-one linear concept-erasure projection inspired by LEACE~\cite{belrose2023leace}, removing from each difference vector the
dominant direction shared across interventions,
\begin{equation}
\Delta h_{\mathrm{erased}} \;=\; (I - P_C)\,\Delta h,
\end{equation}
where $P_C = cc^\top$ and $c$ is the normalized mean intervention direction.
This operation exactly removes the designated shared direction from every
transformed vector. The removed component represents the presence of an
intervention, rather than the scientific problem or any named cognitive label.
The remaining structure therefore represents variation in \emph{how} the
model was asked to reason, rather than simply \emph{that} it was asked to
change. Figure~\ref{fig:residual} is computed entirely on post-erasure
directions.

Figure~\ref{fig:residual} measures how strongly a direction written at source
layer $i$ aligns with what later layer $j$ can read. Each cell of the
$61 \times 61$ matrix is the cosine between the mean post-erasure directions at
the two layers,
\begin{equation}
A_{ij} \;=\; \cos\!\bigl(u^{(i)},\, v^{(j)}\bigr) \;=\;
\frac{\langle u^{(i)},\, v^{(j)} \rangle}{\|u^{(i)}\|\,\|v^{(j)}\|},
\end{equation}
\noindent restricted to $j \geq i$ since we claim only forward propagation. If
cognitive control were diffuse or superficial, alignment mass would spread
broadly across this matrix. Instead, it contains pronounced horizontal bands:
rows 36 through 38 and row 48 concentrate a disproportionate share of total
alignment mass relative to the uniform expectation over 61 source layers, and
their mean outgoing cosine exceeds the 95th percentile of a permutation null
that sign-flips and re-pairs difference vectors. Layer 36
begins the principal write band, layer 38 strengthens it, and layer 48 captures
a late hotspot before output. Earlier layers appear to interpret the context,
while these mid-depth layers establish a more persistent reasoning
configuration.

Residual alignment locates a configuration in the residual stream; it does
not say how later layers read it. A second analysis therefore uses only the
frozen attention weights, with neither activations nor judgment data. Under
MLA, layer~$i$ writes through
$W_O^{(i)}\in\mathbb{R}^{7168\times 8192}$, and layer~$j$ reads
query-relevant residual directions through the down-projection
$W_{Qa}^{(j)}\in\mathbb{R}^{1536\times 7168}$. The write subspace is the
top-$r$ left singular vectors of $W_O^{(i)}$; the read subspace is the
top-$r$ right singular vectors of $W_{Qa}^{(j)}$. Alignment follows the
compositional subspace statistic of~\cite{song2025composition}:
\begin{equation}
B_{ij} \;=\; \sigma_{\max}\!\bigl(U^{(i)\top} V^{(j)}\bigr),
\end{equation}
\noindent where $U^{(i)},V^{(j)}\in\mathbb{R}^{7168\times r}$ are the
orthonormal write and read bases. We use $r=50$. The null is $200$
independent pairs of Haar-random $r$-planes in $\mathbb{R}^{7168}$, each
scored with the same statistic; its 95th percentile is $0.170$.
Figure~\ref{fig:attention} is this write-to-query map on the same 61-layer
profile as Figure~\ref{fig:residual}. Observed $B_{ij}$ on the mid-depth
bands is several times the null. Layers 36 and 38 again appear as a write
hub into the same late region identified by the activation analysis.

The two results are complementary rather than independent: they probe the same
weights through different observables, scientist-conditioned activations after
erasure in one case, frozen attention maps in the other. The complementarity
matters because Figure~\ref{fig:attention} never touches the judgment pairs, so
convergence on the same depth regime cannot be an artifact of looking only in
our own data. The two maps also resolve different components: layer 42 appears
most clearly in the attention pathways, making it the attention-side read hub,
whereas layer 48 is strongest in the residual-stream structure, making it the
late residual peak. Metacognitive Steering therefore targets the intersection
of the two maps: layers 36 and 38 establishing the reasoning configuration,
layer 42 coordinating how attention reads it, and layer 48 carrying the
strongest late activation signature before output. These analyses establish
where steering-relevant information flows; the claim that these layers form a
functional multi-layer reasoning circuit is causal, and is tested by the
steering results below.

\subsection{Scientist Interaction Data Contains Coordinated Cross-Layer Structure}

A useful steering signal should not appear as unrelated noise at each layer. If
the contrastive interventions capture coherent changes in cognition, the same
underlying variation should be expressed across multiple depths.

\begin{figure}[H]
  \centering
  \includegraphics[width=\textwidth]{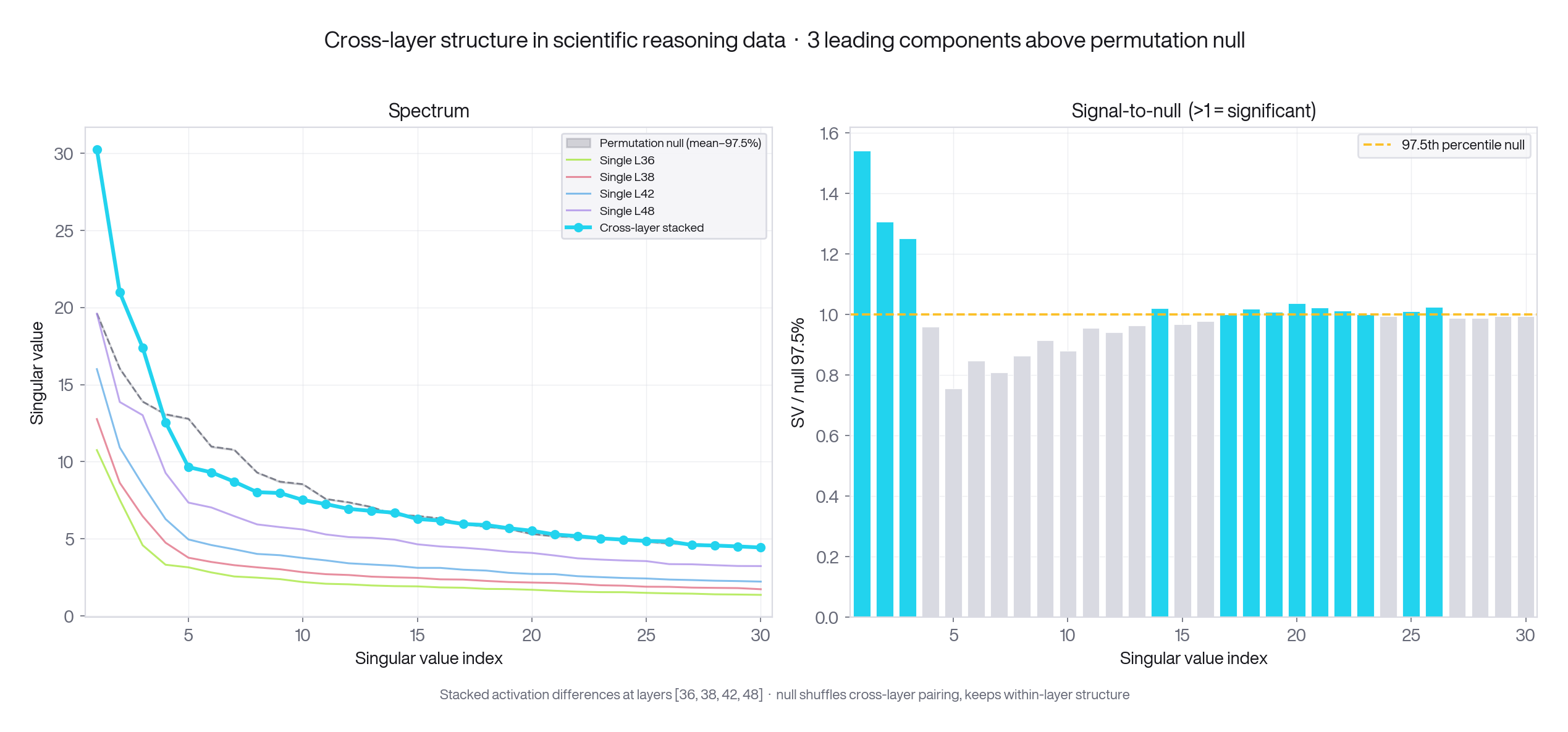}
  \caption{Cross-layer singular value decomposition of concatenated
post-erasure activation differences at layers 36, 38, 42, and 48.
The left panel compares the stacked cross-layer spectrum with spectra
computed at individual layers. The right panel compares the observed
components with an independently shuffled cross-layer permutation null;
the three leading components exceed its 97.5th percentile.}
  \label{fig:svd}
\end{figure}

Figure~\ref{fig:svd} tests this by stacking activation differences from layers
36, 38, 42, and 48 (the candidate control surface identified in
Section 3.1) and applying singular value decomposition, the
standard tool for exposing the correlation structure of network
representations~\cite{bermeitinger2019svd}. Concretely, each contrastive pair
contributes one row to a matrix $X$ formed by concatenating the post-erasure
difference vectors across the four layers,
\begin{equation}
X \;=\; \bigl[\, \Delta h^{(36)} \;\big\|\; \Delta h^{(38)} \;\big\|\;
\Delta h^{(42)} \;\big\|\; \Delta h^{(48)} \,\bigr] \;\in\;
\mathbb{R}^{N_{\text{pairs}} \times 4d}
\end{equation}
with $d = 7168$ and each layer block normalized to unit mean norm so that no
layer dominates the spectrum by scale alone. The observed spectrum is compared
with a permutation null that preserves each layer's internal structure while
destroying the correspondence between the same contrastive pair across layers:
within each layer block, pair identities are shuffled independently over 1{,}000
draws, and the test statistic is recomputed on each draw. This is the standard
permutation construction, in which the null distribution of a statistic is
obtained by recomputing it over rearrangements of the data that erase the
association under test~\cite{paschali2022permutation}.

Three leading components exceed the 97.5th percentile of the permutation null
(one-sided), with signal-to-null ratios of approximately 1.55, 1.3, and 1.25
(Figure~\ref{fig:svd}, right). Comparing an observed spectrum against the
spectrum generated by structureless data is the standard criterion for
separating signal from random bulk~\cite{martin2021rmt}. The stacked spectrum
also exceeds the spectrum computed at any single layer (Figure~\ref{fig:svd},
left), indicating coordination across depths rather than inheritance from one
strong layer. The SVD result answers a foundational question: does the dataset
contain reproducible, coordinated variation spanning multiple layers?

Scientist interventions are not producing independent stylistic perturbations
at different depths. They generate low-dimensional structure that remains
coupled as computation moves through the model. The signal has enough
consistency to survive a stringent cross-layer pairing test.

Together with the lexical controls of Section 3.1, this is
evidence that the dataset is capturing latent organization in reasoning, not
just vocabulary or surface instruction following.

\subsection{A Meta-Cognitive Controller for Long-Horizon Research}

Activation Addition,
Contrastive Activation Addition, and Representation Engineering established
that frozen models can be controlled through interventions in their internal
representations
\cite{turner2023activation,rimsky2024caa,zou2023representation}. More recent
methods make these interventions conditional on input context, inferred model
state, or local representation geometry. MCS extends this approach to
long-horizon scientific reasoning, routing among interventions for exploration,
procedural convergence, and critical reassessment. As shown in
Figure~\ref{fig:param-scale}, previous dynamic steering demonstrations have
largely used models between 7B and 70B parameters. We test MCS on Kimi~2.6
\cite{moonshot2026kimi}, extending the demonstrated scale to a
trillion-parameter mixture-of-experts model, with approximately 32B parameters
activated per token.

\setcounter{figure}{3}
\begin{figure}[H]
  \centering
  \includegraphics[
    width=0.74\textwidth,
    trim=2mm 0mm 0mm 2mm,
    clip
  ]{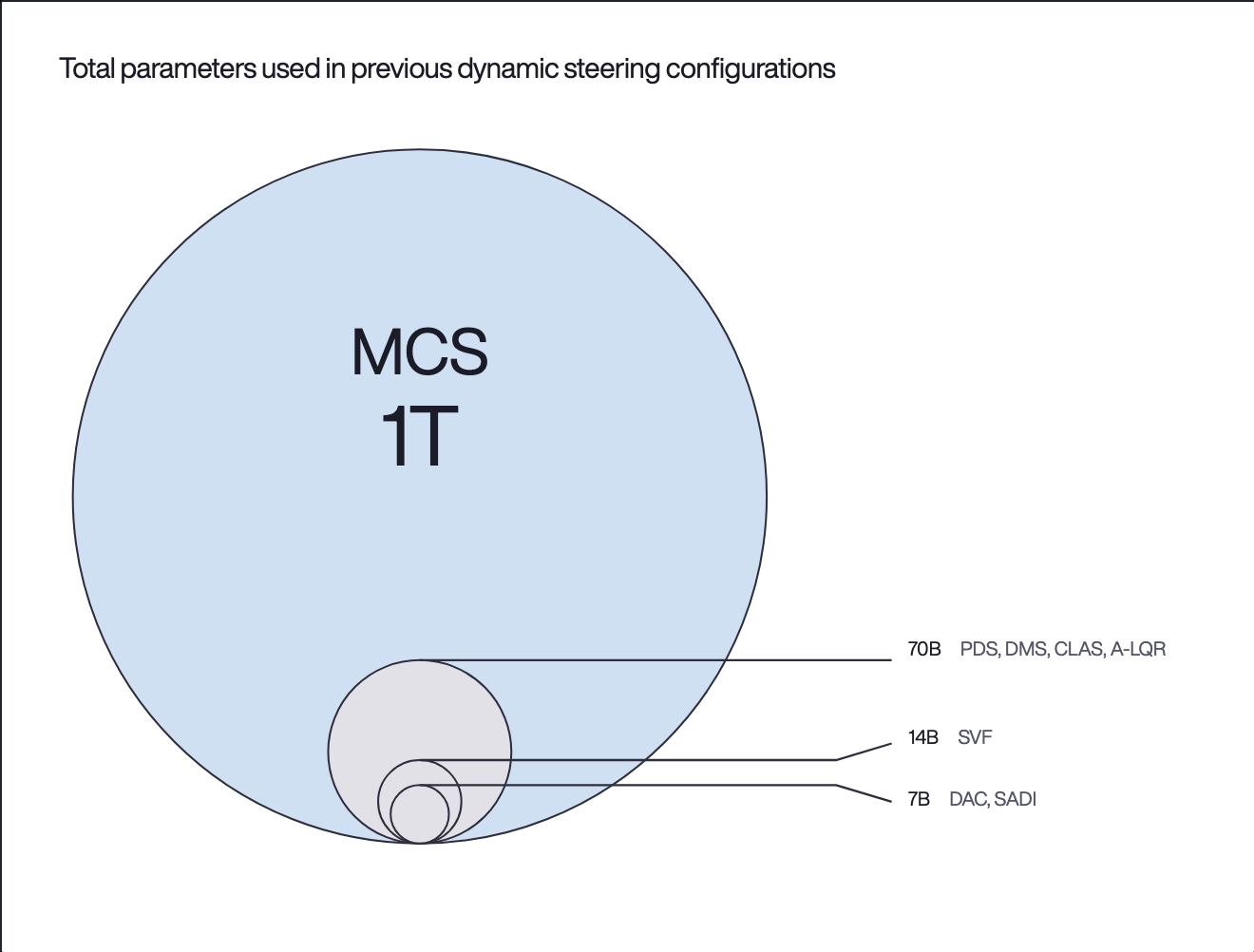}
  \caption{MCS compared with representative dynamic activation-steering
  configurations: PDS \cite{kayan2025pds}, DMS \cite{zhang2025dms},
  CLAS \cite{hsu2026clas}, A-LQR \cite{skifstad2026local},
  SVF \cite{li2026svf}, Dynamic Activation Composition
  \cite{scalena2024dynamic}, and SADI \cite{wang2025sadi}.
  Circle area represents the total parameter count of the largest model
  configuration reported for each method. For MCS, 1T denotes total
  parameters; approximately 32B parameters are activated per token.}
  \label{fig:param-scale}
\end{figure}

Our contrastive scientist-interaction data reveal a compact structure for this
control problem. We analyze two objects derived independently from matched
scientific contexts: un-differenced residual activations at layer 36, which
represent the research context formed by the model before steering, and
contrastive differences at layer 48, which describe directions along which
that state can be shifted. For state discovery, we standardize the layer-36
activations, project them onto 50 principal components, and apply $k$-means
with multiple random initializations. We select the number of states by
silhouette score over $k\in\{3,\ldots,10\}$. The three-state solution scores
$0.457$, compared with $0.304$ for the next-best tested value.

For steering-mode discovery, we take layer-48 contrastive
differences by Euclidean norm, normalize them to unit length, project them onto
50 principal components, and cluster the resulting directions independently.
The procedure initially identifies six modes; one diffuse mode is excluded
from the state-to-mode analysis, leaving five well-resolved intervention modes.
Neither state nor mode labels are supplied to either clustering procedure.
Scientists assigned the functional labels only afterward, by examining the
associated reasoning traces and the effects of controlled intervention.

\begin{table}[H]
  \centering
  \small
  \renewcommand{\arraystretch}{1.25}
  \begin{tabular}{@{}p{0.20\textwidth}p{0.49\textwidth}p{0.25\textwidth}@{}}
    \toprule
    \textbf{Control regime}
    & \textbf{Functional behavior}
    & \textbf{Centroid geometry} \\
    \midrule
    \textbf{Deliberative}\\[-0.2em]
    \emph{Explore}
    & Broad search, speculative branching, and resistance to premature
      commitment
    & Primary reasoning manifold \\
    \addlinespace
    \textbf{Methodical}\\[-0.2em]
    \emph{Converge}
    & Controls, metrics, falsification criteria, and budget-aware execution
    & Approximately $85^\circ$ from deliberative; nearly orthogonal \\
    \addlinespace
    \textbf{Challenging}\\[-0.2em]
    \emph{Scrutinize}
    & Constraint enforcement, error detection, and reversal after
      overextension
    & $130^\circ$--$142^\circ$ from the other regimes; directionally opposed \\
    \bottomrule
  \end{tabular}
  \caption{Activation-defined control regimes. Angles are measured between
  normalized regime-centroid directions in the state representation space.}
  \label{tab:control-regimes}
\end{table}

\begin{figure}[H]
  \centering
  \includegraphics[width=\textwidth]{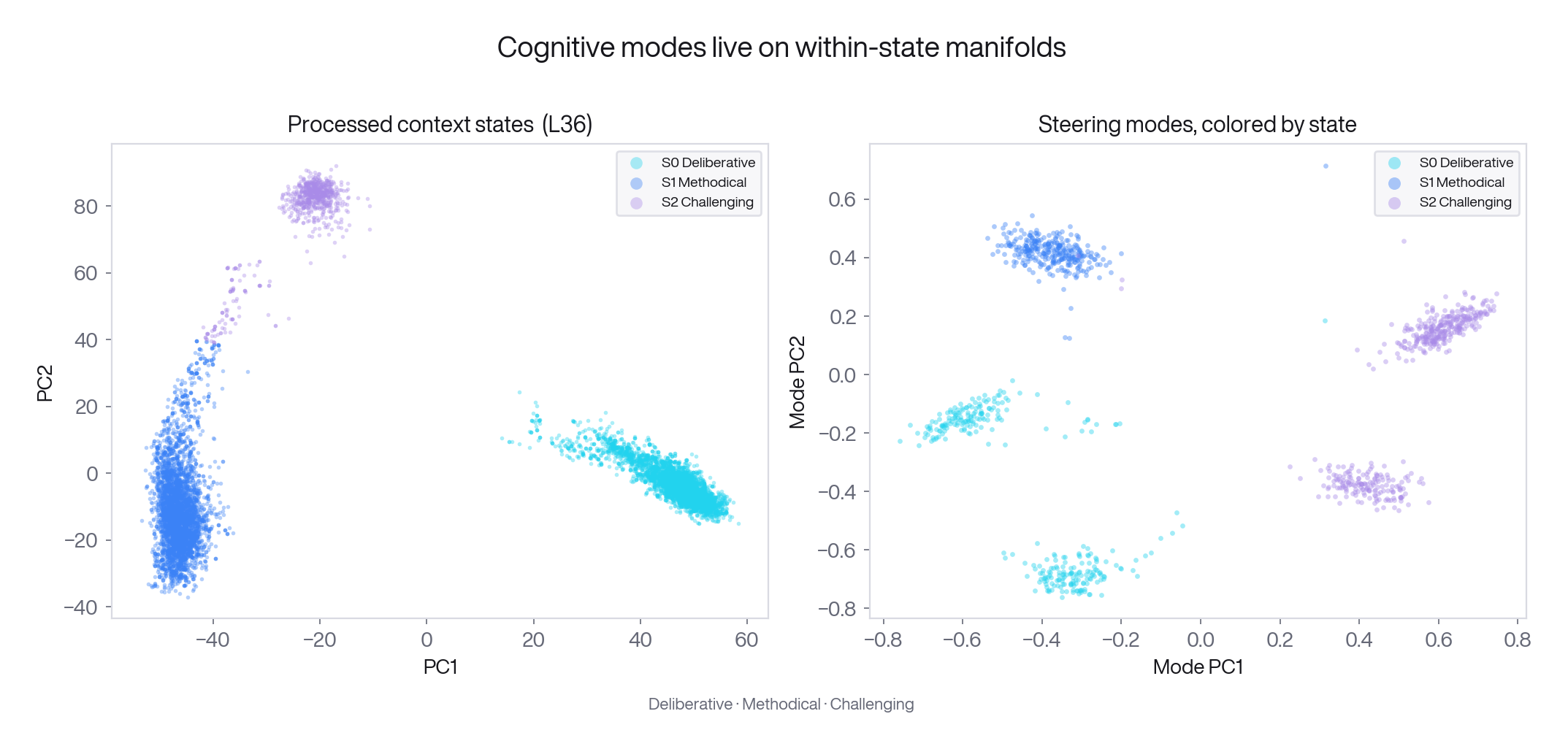}
  \caption{Cognitive states, steering modes, and their conditional mapping.
  Layer-36 un-differenced activations form three state regimes, while
  layer-48 contrastive differences form finer intervention modes. The two
  spaces are clustered independently; color shows their subsequent
  correspondence.}
  \label{fig:state-modes}
\end{figure}

The geometry motivates the routing policy. Methodical and deliberative states
are nearly orthogonal, while challenging points against both. The controller
therefore switches discretely between regimes and blends only between
compatible modes within a regime. Challenging acts as a computational brake
rather than another variety of deliberation. Figure~\ref{fig:state-modes}
summarizes the empirical mapping, which is correspondingly sharp: methodical
states map almost entirely to one intervention, while deliberative and
challenging states each divide into two modes. On the retained high-signal
subset, the association between independently clustered state and mode
assignments reaches Cram\'er's $V=0.993$. This near-deterministic relationship
provides strong evidence that state and steering direction are organized as a
common control system rather than paired arbitrarily.

\begin{figure}[H]
  \centering
  \includegraphics[width=\textwidth]{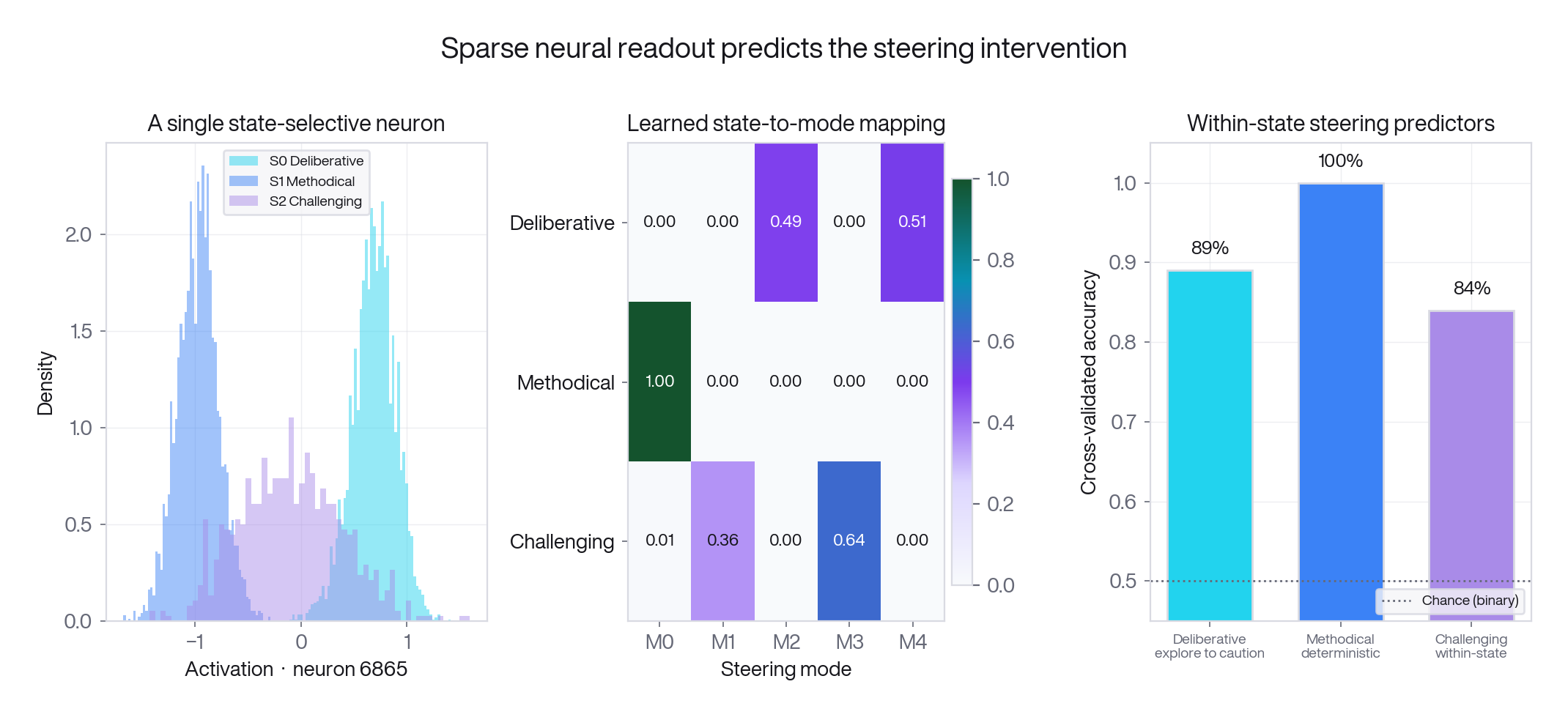}
  \caption{Sparse state detection and dynamic steering composition. Layer-36
  neurons identify the current regime, state-specific predictors assign
  probabilities to compatible modes, and those probabilities blend fixed
  residual directions at layers 36, 38, 42, and 48.}
  \label{fig:neuron-controller}
\end{figure}

Figure~\ref{fig:neuron-controller} shows that these states are accessible
through a surprisingly sparse neural signal, consistent with prior evidence
that high-level contextual and reasoning features can be localized using
sparse probes \cite{gurnee2023finding,dong2026neurons}. In an exploratory
five-fold analysis, ten selected neurons recover the three activation-defined
regimes with approximately 99\% accuracy. This result establishes sparse
recoverability of the state geometry. The operational controller uses a
broader, stability-selected readout together with a dedicated interrupt
detector, favoring robustness across changing research contexts over the
smallest possible probe.

At inference time, an unsteered readout captures the layer-36 residual at the
first generated-token position. A compact neural controller identifies the
current regime, after which small state-specific predictors assign probabilities
to the compatible steering modes. Methodical routing selects one mode
deterministically; deliberative and challenging routing interpolate between
two modes. The internal predictor architectures and training procedure are
held fixed across the experiments reported here.

Let $s(x)$ denote the selected state and $\mathcal{C}_{s(x)}$ its compatible
modes. Given mode probabilities $p(c\mid x,s)$, the intervention at layer
$\ell$ is
\begin{equation}
v_{\ell}(x)
  \;=\;
  \sum_{c\in\mathcal{C}_{s(x)}}
  p(c\mid x,s)\,v_{c,\ell},
\end{equation}
\noindent where $v_{c,\ell}\in\mathbb{R}^{7168}$ is the fixed contrastive
direction for mode $c$ at layer $\ell$. The four layer-specific directions form
the steering matrix
\begin{equation}
S(x)
  \;=\;
  \begin{bmatrix}
    v_{36}(x)^{\top} \\
    v_{38}(x)^{\top} \\
    v_{42}(x)^{\top} \\
    v_{48}(x)^{\top}
  \end{bmatrix}
  \in\mathbb{R}^{4\times7168}.
\end{equation}
\noindent The controller therefore predicts a state and a mode mixture; the
matrix is assembled from empirically derived layer-wise directions rather than
predicted coordinate by coordinate.

During generation, the residual update is
\begin{equation}
h_{t}^{(\ell)}
  \;\leftarrow\;
h_{t}^{(\ell)}+\alpha(x)\,v_{\ell}(x),
\qquad
\ell\in\{36,38,42,48\}.
\end{equation}
The steering strength depends on the selected modes and is damped when the
state readout lies outside the training distribution. Routing occurs once per
response, while the resulting vectors are applied during decoding before the
final layers produce logits. No gradient is computed and no model parameter
changes.

The two-dimensional PCA views in Figure~\ref{fig:state-modes} are used only for
visualization; state and mode discovery operate in their respective
50-component representations. Taken together, Figures~\ref{fig:state-modes}
and~\ref{fig:neuron-controller} suggest that long-horizon scientific cognition
occupies a structured control space that can be detected sparsely and steered
dynamically. Scientist interaction data may therefore provide supervision for
a policy over internal computation itself: when to remain open, when to commit
to method, and when to stop and challenge the path being taken.

\section{Behavioral Analysis}

To observe behavioral differences elicited by Metacognitive Steering, we gave the model two research tasks with different tool sets and analyzed its behavior against the baseline Kimi 2.6.

\begin{figure}[H]
  \centering
  \includegraphics[width=\textwidth]{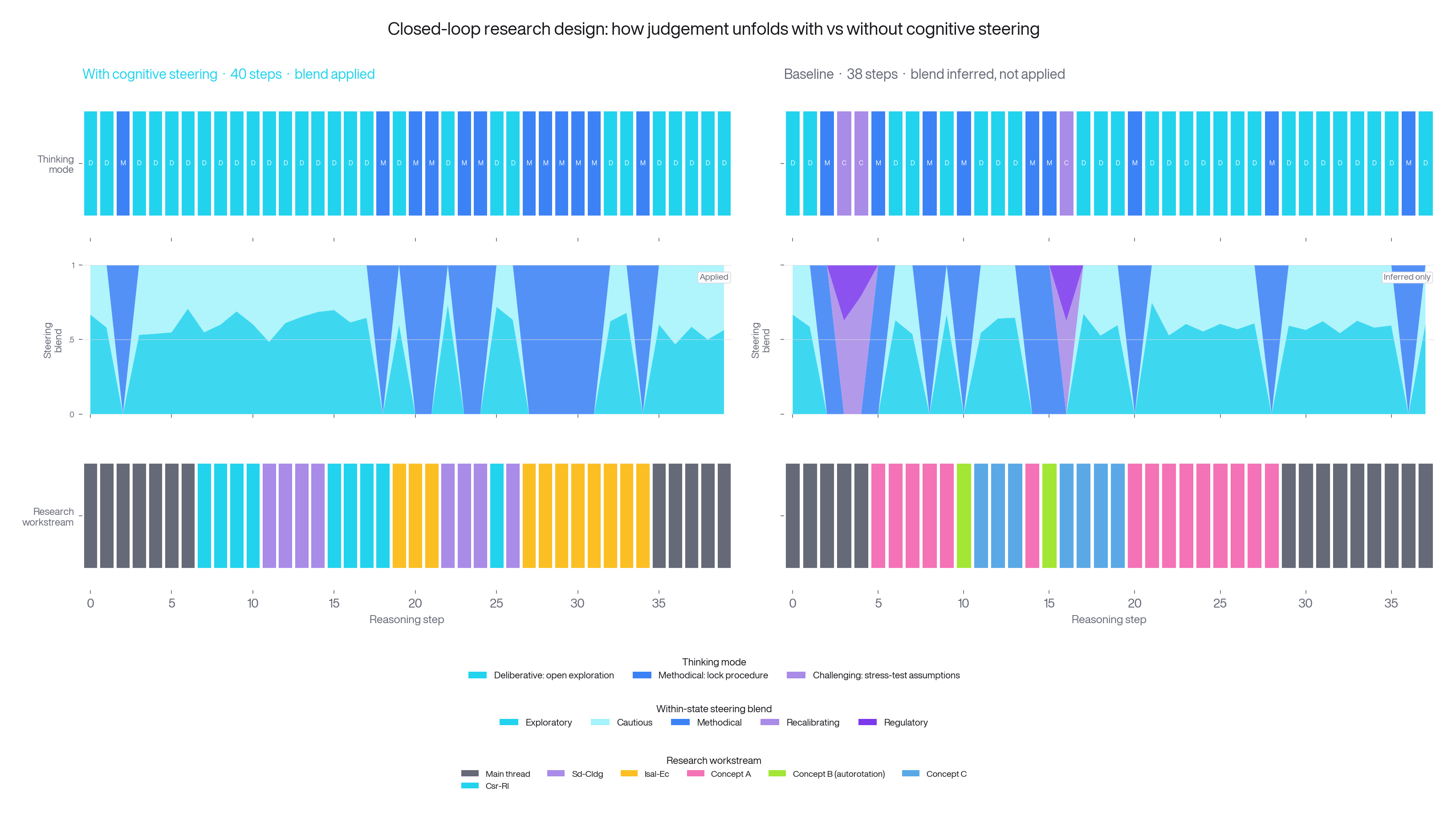}
  \caption{Baseline and Metacognitive Steering trajectories on a
closed-loop inference-time research task. The comparison aligns cognitive
state, steering mixture, and active research workstream across the
investigation. The steered trajectory sustains exploration later into the
campaign, revisits competing approaches, and explicitly prunes a developed
proposal before synthesis.}
  \label{fig:closed-loop-trajectory}
\end{figure}

Figure~\ref{fig:closed-loop-trajectory} shows a detailed trajectory comparison with cognitive state, steering blend, and research workstream side by side on a closed-loop inference-time research problem, in which we gave the model access to web search, code execution, and note taking tools.

The baseline produces competent work, but its exploration is front-loaded. It briefly samples several branches, spends only a few steps on some alternatives, then commits to one path for most of the remaining investigation. It continues searching and cataloguing risks, but does not explicitly prune a developed approach before synthesis.

The steered run develops a different shape. It switches between workstreams slightly more overall and, more importantly, continues switching later in the campaign. It explores more, searches less, and repeatedly returns to competing approaches as new evidence changes the comparison.

It then spends a sustained period examining one approach in depth. During this phase, methodical state becomes concentrated as the agent operationalizes the proposal and tests its failure modes. The approach is ultimately pruned because directly optimizing a sparse state detector could invite representation gaming and destabilize the underlying model. The surviving path is then reframed from first principles and merged into a committed proposal with explicit falsification criteria.

This is closer to scientific judgment because science advances through elimination as much as generation. Producing several plausible ideas is useful. Investigating one seriously, deciding that it should be discarded, and preserving the evidence for why is scientific judgment.

The comparison does not prove that every steered conclusion is correct. It shows that cognitive control changes the process in the intended direction: sustained exploration, deeper testing, explicit pruning, and more decisive synthesis.

\begin{figure}[H]
  \centering
  \includegraphics[width=\textwidth]{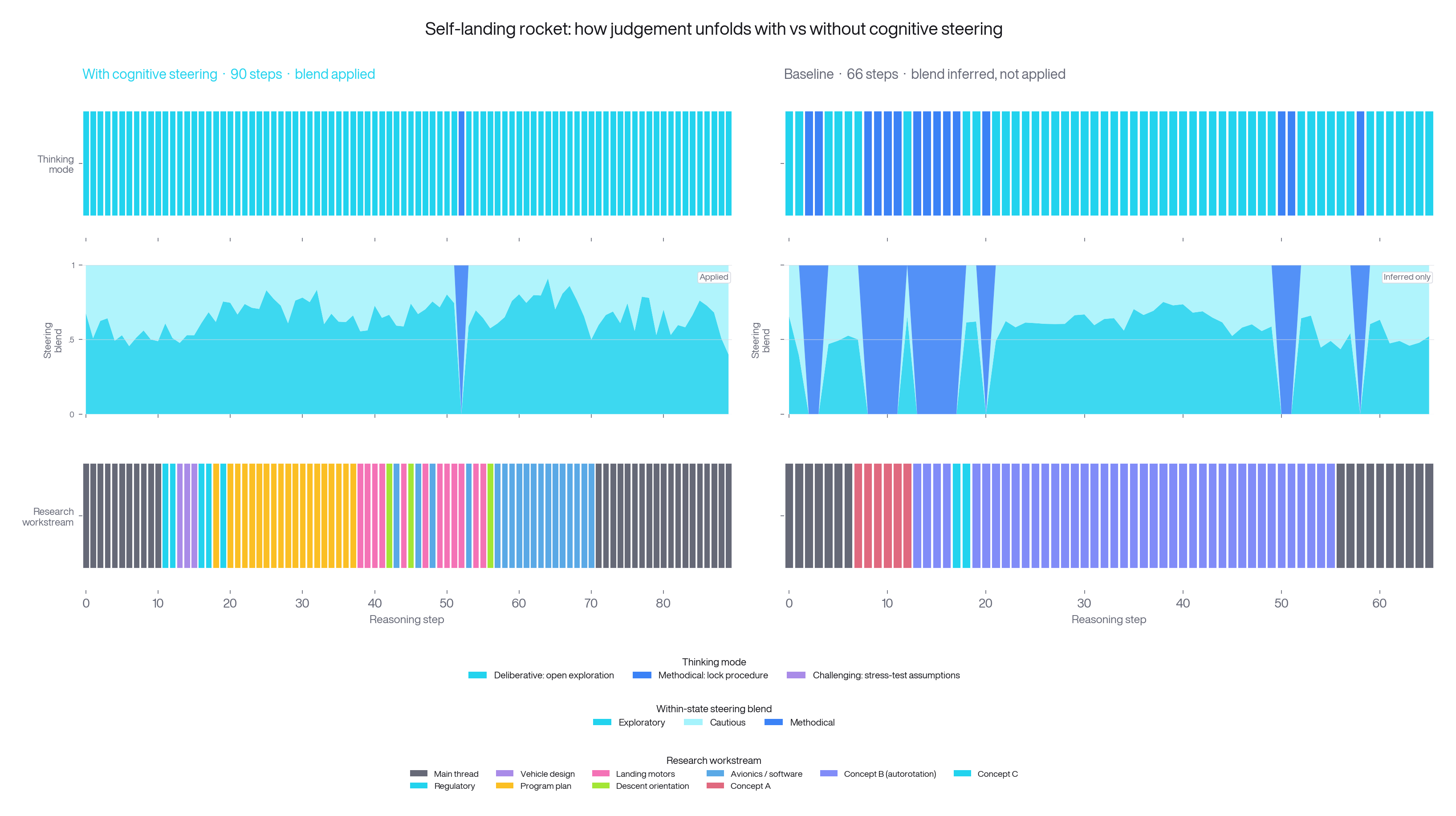}
  \caption{Baseline and Metacognitive Steering trajectories for the design
of a propulsively landing solid-motor rocket. The baseline commits early
to a single recovery concept, whereas the steered trajectory maintains
deliberative exploration across propulsion, avionics, vehicle design,
descent control, and program constraints.}
  \label{fig:rocket-trajectory}
\end{figure}

Figure~\ref{fig:rocket-trajectory} shows another research trajectory in which the model was told to design a rocket that propulsively lands itself using only solid rocket motors. The central difficulty is fundamental: solid rocket motors cannot be throttled. Their thrust and burn profiles are largely fixed once ignited, making controlled landing exceptionally difficult for a vehicle of this size.

The baseline behaves like a capable engineer converging on a mechanism. It briefly opens several concepts, enters methodical mode repeatedly, then spends most of the trajectory developing one branch: autorotative recovery. The result is productive and concrete, but comparatively linear. It selects a plausible mechanism early and executes it.

The steered trajectory remains almost entirely in the deliberative state. On a familiar problem, that could indicate indecision. Here it is appropriate. No established architecture cleanly satisfies the combined propulsion, descent, stability, regulatory, and build constraints. The system therefore keeps the design space open while maintaining separate workstreams for landing motors, descent orientation, avionics, vehicle design, certification, and program planning.

As in the previous trajectory, the difference is not simply more steps. The steered run continues switching between workstreams later in the campaign, revisiting decisions as constraints propagate between domains. It briefly enters methodical mode when a procedure must be fixed, then returns to integration.

A rocket design can be locally correct and globally impossible. The baseline pursues one solution. The steered system maintains a model of the entire program, preserving exploration until the coupled constraints justify commitment.

\section{Real-World Validations}
\subsection{Columbus-1}

To test Metacognitive Steering in the real world, we developed Columbus-1, an
autonomous research system built around the steering approach. We trained a 32B 
bias policy, using the same judgment
dataset. It oversees hierarchical planning and exploration and injects
strategic nudges into the context window of the main reasoning model, similarly
to a scientist steering Aristotle. Columbus-1 maintains persistent memory
through a semantic memory graph that tracks hypotheses, evidence, decisions,
learnings, methods, and facts. It uses this graph to resolve contradictions,
reflect on hypotheses, and generate new hypotheses from causal links and graph
structure.

Figure~\ref{fig:Columbus-1-architecture} summarizes the system architecture.
Goals provide soft and hard constraints to the bias policy, which exchanges
strategic nudges and research context with the steered model. Both components
read from and write to persistent memory, while the steered model uses the
available tool suite to search, write code, analyze data, develop CAD models,
run simulations, and consult human experts.

To explore what our new scientific judgment distillation process could mean for
scientific discovery, we applied it to two distinct and challenging problems
across domains with very different standards of proof. In cybersecurity,
success required finding previously unknown vulnerabilities in production
infrastructure and having an independent laboratory reproduce them. In
aerospace, success required converting a seemingly contradictory propulsion
constraint into a complete, manufacturable vehicle and a controller that
survives realistic simulation. The results offer an early glimpse of the
potential of this paradigm and what it could become at scale.

\setcounter{figure}{8}
\begin{figure}[H]
  \centering
  \includegraphics[width=\textwidth]{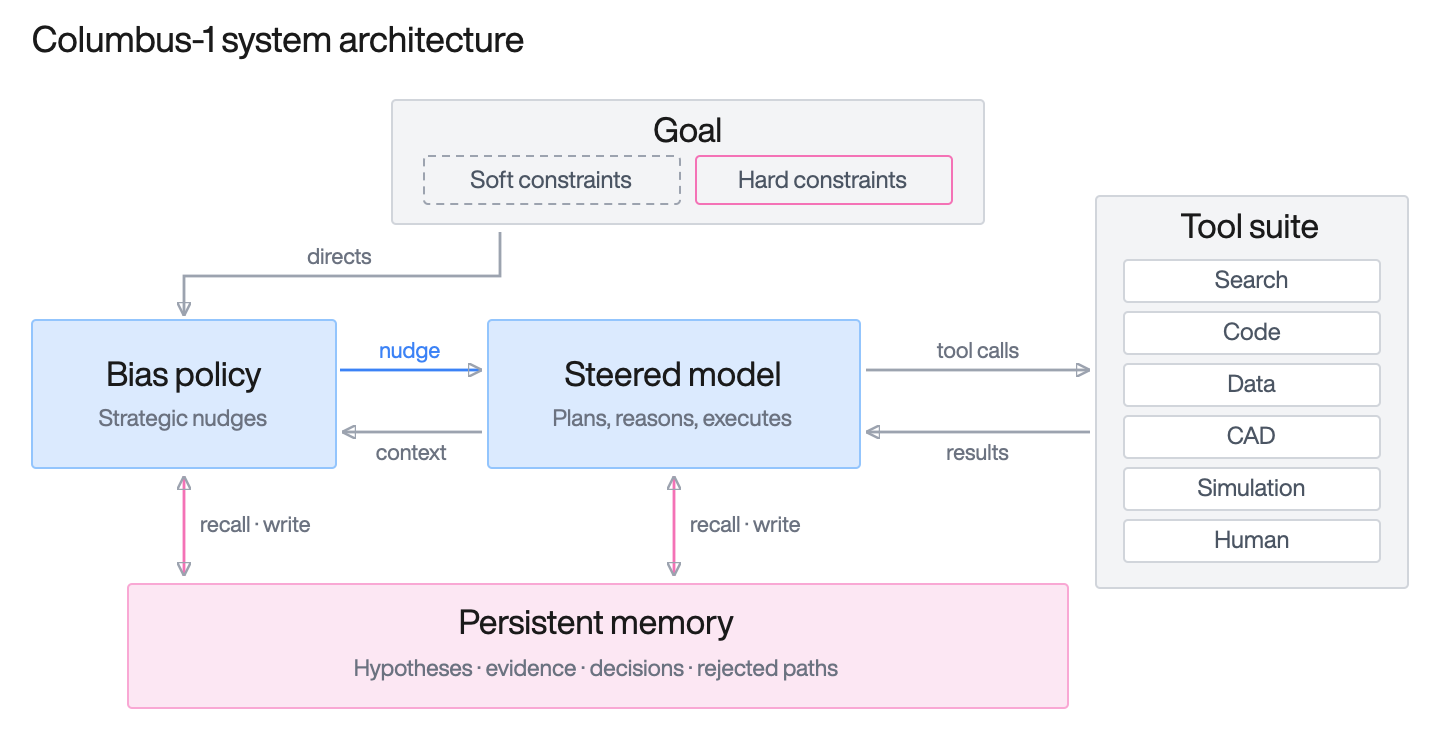}
  \caption{Columbus-1 system architecture. Goals provide soft and hard
constraints to a learned bias policy, which exchanges strategic nudges
and research context with the steered reasoning model. Both components
interact with persistent semantic memory, while the reasoning model uses
tools and human collaborators to search, analyze, simulate, design, and
execute research tasks.}
  \label{fig:Columbus-1-architecture}
\end{figure}

\subsection{Metacognitive Steering in Cyber Discovery}

Discussions with \href{https://xchglabs.com/index.html}{{\textcolor{blue}{\underline{Xchg Labs}}}}, a New York City-based cybersecurity research lab, revealed a fundamental limitation in AI for cybersecurity: models are mode collapsing on the same attack paths and repeatedly find the same classes of vulnerabilities. By nature, reinforcement learning converges on strong but conservative reasoning paths. This is what makes it such a valuable training method for enterprise use cases. But when rewards are ambiguous and the strongest paths often begin as weak signals, this narrows the search space with the number of plausible trajectories growing exponentially with every step.

Finding vulnerabilities that other systems miss requires a model that can preserve competing hypotheses, revisit neglected evidence, and sustain calibrated judgment across investigations lasting days or weeks.

We decided to put Columbus-1 to the test and looked for a problem that would affect millions of people's lives. Bluetooth offered the perfect test case. Few technologies have a greater global reach, having shipped in more than 5.4 billion devices in 2025. Within this enormous ecosystem, BlueZ is the official Linux  Bluetooth protocol stack, deployed across computers, embedded products, industrial infrastructure, and systems operating in healthcare environments, including Linux gateways that communicate with patient monitors, pulse oximeters, and glucose and blood-pressure devices.

The difficulty is not just the size of the codebase. Vulnerability research is a multiplying search tree. Every suspicious copy, parser and state transition branches again at reachability, trust boundaries, allocator behavior, build configuration and exploit mitigations. Most paths fail only after expensive validation. The number of plausible chains grows at every step, while the signal that matters can begin as one unremarkable line of C.

Columbus-1 identified ten crash sites corresponding to eight confirmed attacker-reachable findings. Xchg Labs independently rebuilt the proofs of concept, reproduced the faults and confirmed the findings against BlueZ 5.86. The set included stack and heap corruption, a filesystem race and allocation-failure defects; one finding crossed the radio boundary with no pairing.

Then Columbus-1 found the one that changes the scale of the result. While scanning nearby devices, BlueZ copied an attacker-controlled advertised name into a fixed 250-byte stack buffer inside the privileged bluetoothd daemon and wrote past its end. The input arrives in Bluetooth advertising data, before authentication, before pairing and before the target has agreed to anything.

The highest-severity zero day is a zero-click, full-chain remote code execution vulnerability. No click. No pairing. No connection request. A nearby attacker can compromise the target over the air without the victim doing anything, then execute code with full access to the compromised system. A reliable chain with that combination of reach, silence and control is a nation-state-tier capability. The victim can be owned before the operating system has a reason to put anything on screen.

Disclosure: Xchg Labs contacted the BlueZ maintainers on 6 July 2026 at 04:16 UTC from contact@xchglabs.com and remains available to help.

\subsection{Metacognitive Steering in Aerospace Engineering}

After the BlueZ success, we gave Columbus-1 a problem that initially appeared physically contradictory: autonomously design and oversee the construction of a ten-foot rocket that launches conventionally and lands propulsively using only commercially available G-class solid motors.

Propulsive landing normally depends on engines that can throttle and shut down. Solid motors can do neither. Once ignited, each motor delivers a largely fixed thrust curve and must burn to completion. Columbus-1 therefore had to land the vehicle by controlling discrete ignition events while simultaneously managing attitude, velocity, aerodynamic forces, wind, ignition uncertainty, structural limits, regulatory constraints, and ground contact. A timing error measured in fractions of a second can turn a landing burn into either a crash or a second launch.

This was not a simulation prompt or a design-assistance exercise. Columbus-1 was given a budget, procurement constraints, a compressed schedule, and responsibility for the complete engineering program. It selected the architecture, generated parametric CAD, produced fabrication-ready STL, STEP, and DXF files, designed motor cassettes, avionics structures, landing-leg attachments, and energy-absorbing hardware, and revised components as measured hardware and procurement information became available.

The fabrication team communicated with Columbus-1 through iMessage. They provided measurements, photographs, manufacturing constraints, and physical observations. Columbus-1 interpreted that feedback, updated the design, and issued the next set of build instructions. The humans operated the printers, assembled components, and performed work that required physical access. Columbus-1 maintained the system-level model and directed the engineering process.

In parallel, Columbus-1 wrote and iteratively improved the flight software. It executed its controllers through an OpenRocket ascent model coupled to a closed-loop six-degree-of-freedom landing simulation incorporating aerodynamics, rigid-body motion, wind, motor dispersion, ignition uncertainty, thrust-vector control, cold-gas attitude control, landing dynamics, and noisy multi-rate sensors.

Being aware of the capabilities of a human judgment enhanced model, we had an employee watch Columbus-1 24/7 to ensure safety but also check the reasoning traces, as observability and reproducibility will be essential parts in the new era of AI led science. A SpaceX propulsion engineer and a PhD avionics expert reviewed the simulation environment, assumptions, and reported results as independent sanity and safety checks. Columbus-1's architecture and control logic survived their review without requiring corrective intervention.

\begin{figure}[H]
  \centering
  \includegraphics[width=0.9\textwidth]{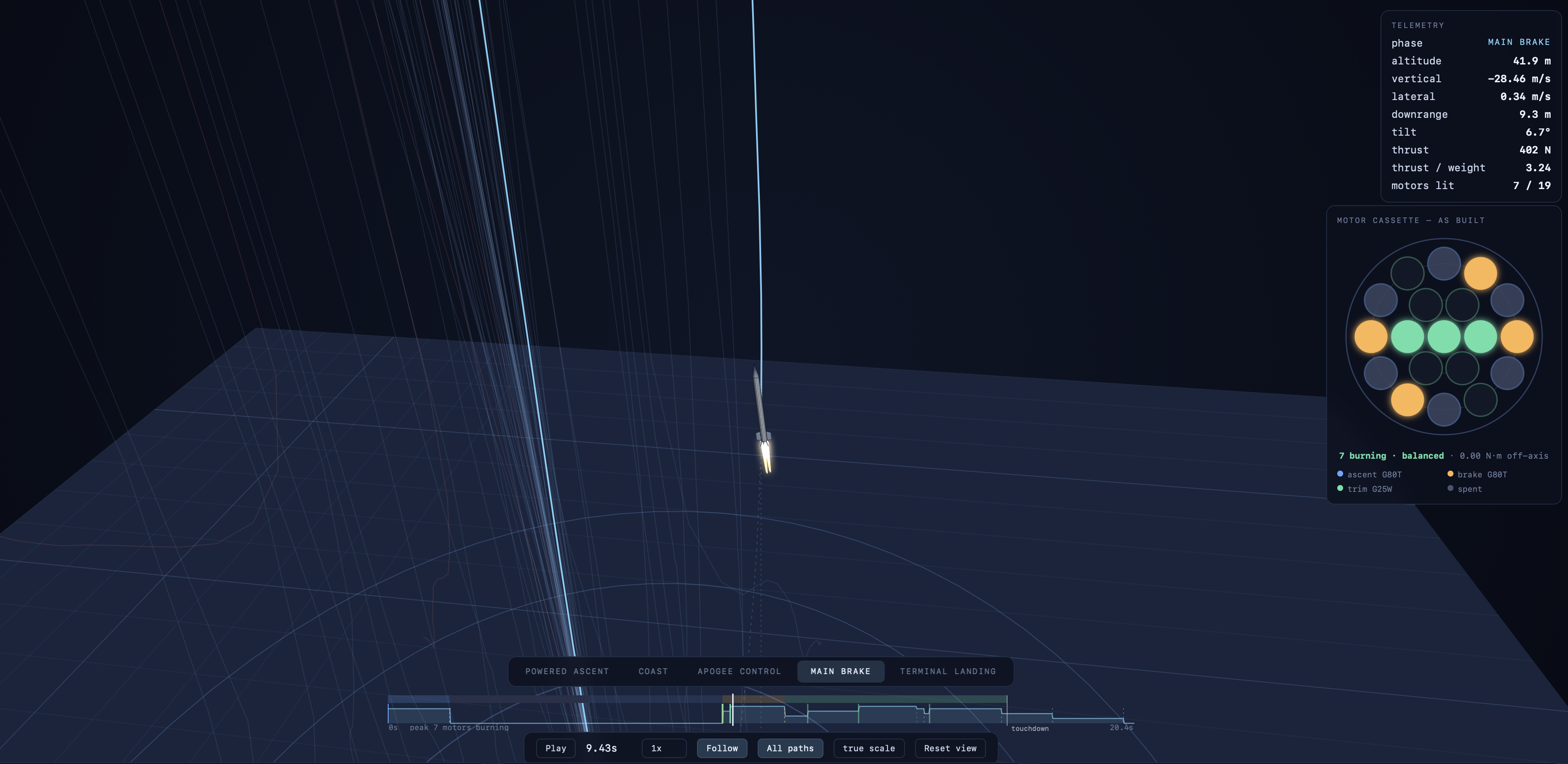}
  \caption{Closed-loop six-degree-of-freedom simulation of the propulsive
landing system. The visualization shows descent trajectory, flight phases,
vehicle telemetry, and sequential solid-motor ignition. Staggered ignition
approximates throttle control by coordinating predicted time-to-contact
with descent velocity and remaining motor impulse.
\href{https://dart1-flight.surge.sh}
{\textcolor{blue}{\underline{View the flight simulation}}}.}
  \label{fig:flight-sim}
\end{figure}

Through autonomous experimentation, Columbus-1 discovered that minimizing touchdown velocity was the wrong objective. A vehicle can reach the ground slowly and still fail if a burning solid motor contains enough residual impulse to launch it again. The controller instead had to optimize arrival velocity and remaining impulse together. Columbus-1 further discovered that attitude must be stabilized early in descent, before terminal propulsion, and that motor staging should be driven by predicted time-to-contact and remaining impulse rather than fixed altitude thresholds. This led to a staggered ignition corridor: the controller tracks a target relationship between altitude and descent velocity and ignites motors sequentially in small increments as the vehicle departs from that corridor. Each irreversible ignition raises thrust by one step, creating a staircase-style approximation of throttling with solid motors that cannot themselves throttle or shut down, as visualized in Figure~\ref{fig:flight-sim}.

These were not supplied design rules. They emerged from failed simulations, competing controller hypotheses, and repeated revision. Columbus-1 converted them into a predictive, phase-aware landing architecture.

Columbus-1 preserved every controller version, telemetry trace, failed experiment, and causal lesson so that each result became evidence for the next design decision.

The physical flight remains the decisive test. But the organizational shift has already occurred.

AI is no longer assisting a human-led engineering program. Columbus-1 is overseeing the program, maintaining the design logic, running the experiments, and directing physical execution through human collaborators. Within weeks, it moved from an apparently incompatible set of constraints to reviewed simulation, manufacturing files, flight software, and hardware under construction.

The transition from AI as a tool to AI as an autonomous research leader is not theoretical. It is currently being fabricated.

\section{Limitations}

This work provides evidence that scientific judgment can be distilled
into a neural controller, but its transfer across models, domains, and research
horizons requires further study. Our dataset is shaped by the scientific fields
and problems currently represented on the platform, leaving some domains
substantially underrepresented. Expanding it into a larger and more balanced
cross-disciplinary corpus is an important next step.

Scientific discovery also lacks robust, standardized benchmarks against which
systems of this kind can be evaluated. Although we developed LLM judges for
research quality and cognitive approach, stronger evaluation will require
greater involvement from domain experts and measures tied to reproducibility
and real scientific outcomes. The rocket architecture and controller have so
far been validated in simulation; completing fabrication and testing the system
under real-world flight conditions is our immediate next step.

\section{Discussion and Future Work}
The results presented here offer a glimpse of a broader possibility:
scientific judgment may be learnable not only as a property of model weights,
but as a policy over internal computation. The cross-layer analyses identify a
structured neural control surface, while the behavioral trajectories show that
intervening on this structure changes when the model explores, converges, and
challenges its own direction. Scientific reasoning therefore appears less like
a single capability to elicit and more like a sequence of cognitive regimes
that must be selected as evidence changes. Process-level scientist interaction
data provide supervision for that selection problem.

A natural extension is a controller that continues learning during research.
The present framework selects interventions from the model's current state, but
future controllers could also learn from the consequences of earlier decisions:
which hypotheses survived experimentation, which searches produced useful
evidence, which commitments were premature, and which reversals improved the
eventual result. Persistent research memory supplies the necessary history.
This would turn metacognitive control from a fixed policy into a bounded online
learning system that improves across experiments while leaving the underlying
reasoning model unchanged. Such a controller could operate across multiple
timescales, making immediate steering decisions within a response while also
adjusting its research strategy across an entire project.

A second direction is scientist-specific metacognitive control. With more
sample-efficient adaptation, a controller could learn from the traces of a
single scientist and reproduce aspects of how that scientist allocates
attention, responds to uncertainty, evaluates weak evidence, and decides when
to abandon an idea. A population-level controller could provide a shared prior,
with a small number of interactions adapting it to an individual researcher.
This would make it possible to study which elements of scientific judgment
transfer across researchers and which reflect domain expertise or individual
style. Controllers could also combine complementary policies from several
scientists rather than reducing scientific judgment to one universal strategy.

The observed low-dimensional structure motivates a narrower question about
cross-model transfer. Specific neurons, layers, and steering directions should
not be expected to transfer directly because model architectures and internal
representations differ. However, the higher-level policy of when to explore,
converge, or scrutinize may be reusable if each model is given its own state
readout and intervention directions. Future work should test whether analogous
control regimes emerge across model families and whether a shared routing
policy can operate through model-specific controllers. Answering this question
will require broader models and disciplines, together with evaluation
environments that measure evidence acquisition, belief revision, experimental
value, and eventual reproducibility. The present work provides an initial
demonstration of the control paradigm rather than evidence of universal
transfer.


\begin{thebibliography}{99}

\bibitem{abdullaev2026chars}
L. U. Abdullaev, N. Y. L. Wong, R. T. Z. Lee, S. Jiang, K. N. M. Nguyen, and T. M. Nguyen,
``Concept heterogeneity-aware representation steering,''
in \textit{Proc. 43rd Int. Conf. Mach. Learn.}, 2026.
[Online]. Available: \url{https://arxiv.org/abs/2603.02237}.

\bibitem{belrose2023leace}
N. Belrose, D. Schneider-Joseph, S. Ravfogel, R. Cotterell, E. Raff, and S. Biderman,
``LEACE: Perfect linear concept erasure in closed form,''
in \textit{Advances in Neural Information Processing Systems},
vol. 36, pp. 66044--66063, 2023.
[Online]. Available: \url{https://proceedings.neurips.cc/paper_files/paper/2023/hash/d066d21c619d0a78c5b557fa3291a8f4-Abstract-Conference.html}.

\bibitem{bermeitinger2019svd}
B. Bermeitinger, S. Handschuh, and T. A. Runkler,
``Singular value decomposition and neural networks,''
\textit{arXiv preprint arXiv:1906.11755}, 2019,
doi: \href{https://doi.org/10.48550/arXiv.1906.11755}{10.48550/arXiv.1906.11755}.
[Online]. Available: \url{https://arxiv.org/abs/1906.11755}.

\bibitem{chen2025act}
M. Chen, R. Sun, T. Pfister, and S. \"O. Ar{\i}k,
``Learning to clarify: Multi-turn conversations with action-based contrastive self-training,''
in \textit{Proc. Int. Conf. Learn. Representations}, 2025.
[Online]. Available: \url{https://openreview.net/forum?id=SIE6VFps9x}.

\bibitem{deepseekai2025r1}
DeepSeek-AI,
``DeepSeek-R1: Incentivizing reasoning capability in LLMs via reinforcement learning,''
\textit{arXiv preprint arXiv:2501.12948}, 2025.
[Online]. Available: \url{https://arxiv.org/abs/2501.12948}.

\bibitem{deepseekai2024v3}
DeepSeek-AI,
``DeepSeek-V3 technical report,''
\textit{arXiv preprint arXiv:2412.19437}, 2024,
doi: \href{https://doi.org/10.48550/arXiv.2412.19437}{10.48550/arXiv.2412.19437}.
[Online]. Available: \url{https://arxiv.org/abs/2412.19437}.

\bibitem{dong2026neurons}
F. Dong, Z. Yan, X. Ge, Z. Xu, M. Zhang, X. Chen, B. He, X. Xin, Z. Chen, and Y. Zhou,
``Identifying and transferring reasoning-critical neurons: Improving LLM inference reliability via activation steering,''
\textit{arXiv preprint arXiv:2601.19847}, 2026,
doi: \href{https://doi.org/10.48550/arXiv.2601.19847}{10.48550/arXiv.2601.19847}.
[Online]. Available: \url{https://arxiv.org/abs/2601.19847}.

\bibitem{felin2024theory}
T. Felin and M. Holweg,
``Theory is all you need: AI, human cognition, and causal reasoning,''
\textit{Strategy Science},
vol. 9, no. 4, pp. 346--371, 2024,
doi: \href{https://doi.org/10.1287/stsc.2024.0189}{10.1287/stsc.2024.0189}.
[Online]. Available: \url{https://pubsonline.informs.org/doi/10.1287/stsc.2024.0189}.

\bibitem{griot2025metacognition}
M. Griot, C. Hemptinne, J. Vanderdonckt, and D. Yuksel,
``Large language models lack essential metacognition for reliable medical reasoning,''
\textit{Nature Communications},
vol. 16, Art. no. 642, 2025,
doi: \href{https://doi.org/10.1038/s41467-024-55628-6}{10.1038/s41467-024-55628-6}.
[Online]. Available: \url{https://www.nature.com/articles/s41467-024-55628-6}.

\bibitem{gurnee2023finding}
W. Gurnee, N. Nanda, M. Pauly, K. Harvey, D. Troitskii, and D. Bertsimas,
``Finding neurons in a haystack: Case studies with sparse probing,''
\textit{Transactions on Machine Learning Research}, 2023.
[Online]. Available: \url{https://openreview.net/forum?id=JYs1R9IMJr}.

\bibitem{hubert2026nature}
T. Hubert, R. Mehta, L. Sartran, \textit{et al.},
``Olympiad-level formal mathematical reasoning with reinforcement learning,''
\textit{Nature},
vol. 651, pp. 607--613, 2026,
doi: \href{https://doi.org/10.1038/s41586-025-09833-y}{10.1038/s41586-025-09833-y}.
[Online]. Available: \url{https://www.nature.com/articles/s41586-025-09833-y}.

\bibitem{hsu2026clas}
B. Hsu, D. Beaglehole, A. Radhakrishnan, and M. Belkin,
``Contextual linear activation steering of language models,''
\textit{arXiv preprint arXiv:2604.24693}, 2026,
doi: \href{https://doi.org/10.48550/arXiv.2604.24693}{10.48550/arXiv.2604.24693}.
[Online]. Available: \url{https://arxiv.org/abs/2604.24693}.

\bibitem{kayan2025pds}
C. E. Kayan and L. Zhang,
``Prototype-based dynamic steering for large language models,''
\textit{arXiv preprint arXiv:2510.05498}, 2025,
doi: \href{https://doi.org/10.48550/arXiv.2510.05498}{10.48550/arXiv.2510.05498}.
[Online]. Available: \url{https://arxiv.org/abs/2510.05498}.

\bibitem{kumbhar2026cogtrl}
S. Kumbhar, S. Mashetty, D. Handa, K. Coutinho, S. S. Ghule, and C. Baral,
``COGTRL: Training LLMs for scientific discovery assistance using cognitive traces via reinforcement learning,''
\textit{arXiv preprint arXiv:2608.30109}, 2026.
[Online]. Available: \url{https://arxiv.org/abs/2608.30109}.

\bibitem{kuhn1962structure}
T. S. Kuhn,
\textit{The Structure of Scientific Revolutions}.
Chicago, IL, USA: Univ. of Chicago Press, 1962.
[Online]. Available: \url{https://press.uchicago.edu/ucp/books/book/chicago/S/bo13179781.html}.

\bibitem{lee2025cast}
B. W. Lee, I. Padhi, K. N. Ramamurthy, E. Miehling, P. Dognin, M. Nagireddy, and A. Dhurandhar,
``Programming refusal with conditional activation steering,''
in \textit{Proc. Int. Conf. Learn. Representations}, 2025.
[Online]. Available: \url{https://arxiv.org/abs/2409.05907}.

\bibitem{li2026svf}
J. Li, Y. Li, and K.-H. Huang,
``Steering vector fields for context-aware inference-time control in large language models,''
\textit{arXiv preprint arXiv:2602.01654}, 2026,
doi: \href{https://doi.org/10.48550/arXiv.2602.01654}{10.48550/arXiv.2602.01654}.
[Online]. Available: \url{https://arxiv.org/abs/2602.01654}.

\bibitem{lin2024pohf}
X. Lin, Z. Dai, A. Verma, S.-K. Ng, P. Jaillet, and B. K. H. Low,
``Prompt optimization with human feedback,''
\textit{arXiv preprint arXiv:2405.17346}, 2024,
doi: \href{https://doi.org/10.48550/arXiv.2405.17346}{10.48550/arXiv.2405.17346}.
[Online]. Available: \url{https://arxiv.org/abs/2405.17346}.

\bibitem{lu2026mind}
Y.-L. Lu, J. Song, C. Zhang, and W. Wang,
``Mind the gap: The divergence between human and LLM-generated tasks,''
in \textit{Proc. AAAI Conf. Artif. Intell.},
vol. 40, no. 3, pp. 1964--1972, 2026,
doi: \href{https://doi.org/10.1609/aaai.v40i3.37177}{10.1609/aaai.v40i3.37177}.
[Online]. Available: \url{https://ojs.aaai.org/index.php/AAAI/article/view/37177}.

\bibitem{martin2021rmt}
C. H. Martin and M. W. Mahoney,
``Implicit self-regularization in deep neural networks: Evidence from random matrix theory and implications for learning,''
\textit{Journal of Machine Learning Research},
vol. 22, no. 165, pp. 1--73, 2021.
[Online]. Available: \url{https://jmlr.org/papers/v22/20-410.html}.

\bibitem{moonshot2026kimi}
Moonshot AI,
``Kimi K2.6,'' Hugging Face model card, 2026.
[Online]. Available: \url{https://huggingface.co/moonshotai/Kimi-K2.6}. Accessed: Sep. 2, 2026.

\bibitem{paschali2022permutation}
M. Paschali, Q. Zhao, E. Adeli, and K. M. Pohl,
``Bridging the gap between deep learning and hypothesis-driven analysis via permutation testing,''
in \textit{Predictive Intelligence in Medicine: 5th Int. Workshop, PRIME 2022, Held in Conjunction with MICCAI 2022},
Lecture Notes in Computer Science, vol. 13564, pp. 13--23, 2022,
doi: \href{https://doi.org/10.1007/978-3-031-16919-9_2}{10.1007/978-3-031-16919-9\_2}.
[Online]. Available: \url{https://link.springer.com/chapter/10.1007/978-3-031-16919-9_2}.

\bibitem{rimsky2024caa}
N. Rimsky, N. Gabrieli, J. Schulz, M. Tong, E. Hubinger, and A. Turner,
``Steering Llama 2 via contrastive activation addition,''
in \textit{Proc. 62nd Annu. Meeting Assoc. Comput. Linguistics},
vol. 1, pp. 15504--15522, 2024,
doi: \href{https://doi.org/10.18653/v1/2024.acl-long.828}{10.18653/v1/2024.acl-long.828}.
[Online]. Available: \url{https://aclanthology.org/2024.acl-long.828/}.

\bibitem{riosgarcia2026scientists}
M. R\'ios-Garc\'ia, N. Alampara, C. Gupta, I. Mandal, S. Mannan, A. A. Aghajani, N. M. A. Krishnan, and K. M. Jablonka,
``AI scientists produce results without reasoning scientifically,''
\textit{arXiv preprint arXiv:2604.18805}, 2026,
doi: \href{https://doi.org/10.48550/arXiv.2604.18805}{10.48550/arXiv.2604.18805}.
[Online]. Available: \url{https://arxiv.org/abs/2604.18805}.

\bibitem{scalena2024dynamic}
D. Scalena, G. Sarti, and M. Nissim,
``Multi-property steering of large language models with dynamic activation composition,''
in \textit{Proc. 7th BlackboxNLP Workshop: Analyzing and Interpreting Neural Networks for NLP},
pp. 577--603, 2024,
doi: \href{https://doi.org/10.18653/v1/2024.blackboxnlp-1.34}{10.18653/v1/2024.blackboxnlp-1.34}.
[Online]. Available: \url{https://aclanthology.org/2024.blackboxnlp-1.34/}.

\bibitem{skifstad2026local}
J. Skifstad, X. A. Yang, and G. Chou,
``Local linearity of LLMs enables activation steering via model-based linear optimal control,''
\textit{arXiv preprint arXiv:2604.19018}, 2026,
doi: \href{https://doi.org/10.48550/arXiv.2604.19018}{10.48550/arXiv.2604.19018}.
[Online]. Available: \url{https://arxiv.org/abs/2604.19018}.

\bibitem{song2025composition}
J. Song, Z. Xu, and Y. Zhong,
``Out-of-distribution generalization via composition: A lens through induction heads in transformers,''
\textit{Proceedings of the National Academy of Sciences},
vol. 122, no. 6, Art. no. e2417182122, 2025,
doi: \href{https://doi.org/10.1073/pnas.2417182122}{10.1073/pnas.2417182122}.
[Online]. Available: \url{https://www.pnas.org/doi/10.1073/pnas.2417182122}.

\bibitem{turner2023activation}
A. M. Turner, L. Thiergart, G. Leech, D. Udell, J. J. Vazquez, U. Mini, and M. MacDiarmid,
``Steering language models with activation engineering,''
\textit{arXiv preprint arXiv:2308.10248}, 2023,
doi: \href{https://doi.org/10.48550/arXiv.2308.10248}{10.48550/arXiv.2308.10248}.
[Online]. Available: \url{https://arxiv.org/abs/2308.10248}.

\bibitem{venhoff2025reasoning}
C. Venhoff, I. Arcuschin, P. Torr, A. Conmy, and N. Nanda,
``Understanding reasoning in thinking language models via steering vectors,''
in \textit{Workshop on Reasoning and Planning for Large Language Models}, 2025.
[Online]. Available: \url{https://arxiv.org/abs/2506.18167}.

\bibitem{wang2025sadi}
W. Wang, J. Yang, and W. Peng,
``Semantics-adaptive activation intervention for LLMs via dynamic steering vectors,''
in \textit{Proc. Int. Conf. Learn. Representations}, 2025.
[Online]. Available: \url{https://openreview.net/forum?id=8WQ7VTfPTl}.

\bibitem{wei2025verifiers}
J. Wei,
``Asymmetry of verification and verifier's rule,'' Jason Wei Blog, Jul. 15, 2025.
[Online]. Available: \url{https://www.jasonwei.net/blog/asymmetry-of-verification-and-verifiers-law}. Accessed: Sep. 2, 2026.

\bibitem{wilie2024belief}
B. Wilie, S. Cahyawijaya, E. Ishii, J. He, and P. Fung,
``Belief revision: The adaptability of large language models reasoning,''
in \textit{Proc. Conf. Empir. Methods Nat. Lang. Process.},
pp. 10480--10496, 2024,
doi: \href{https://doi.org/10.18653/v1/2024.emnlp-main.586}{10.18653/v1/2024.emnlp-main.586}.
[Online]. Available: \url{https://aclanthology.org/2024.emnlp-main.586/}.

\bibitem{yin2025scientific}
M. Yin, Y. Qu, L. Yang, L. Cong, and M. Wang,
``Toward scientific reasoning in LLMs: Training from expert discussions via reinforcement learning,''
\textit{arXiv preprint arXiv:2505.19501}, 2025,
doi: \href{https://doi.org/10.48550/arXiv.2505.19501}{10.48550/arXiv.2505.19501}.
[Online]. Available: \url{https://arxiv.org/abs/2505.19501}.

\bibitem{zhang2025dms}
X. Zhang,
``Generalization or memorization: Dynamic decoding for mode steering,''
\textit{arXiv preprint arXiv:2510.22099}, 2025,
doi: \href{https://doi.org/10.48550/arXiv.2510.22099}{10.48550/arXiv.2510.22099}.
[Online]. Available: \url{https://arxiv.org/abs/2510.22099}.

\bibitem{ziems2024llms}
C. Ziems, W. Held, O. Shaikh, J. Chen, Z. Zhang, and D. Yang,
``Can large language models transform computational social science?''
\textit{Computational Linguistics},
vol. 50, no. 1, pp. 237--291, 2024,
doi: \href{https://doi.org/10.1162/coli_a_00502}{10.1162/coli\_a\_00502}.
[Online]. Available: \url{https://aclanthology.org/2024.cl-1.8/}.

\bibitem{zou2023representation}
A. Zou, L. Phan, S. Chen, J. Campbell, P. Guo, R. Ren, A. Pan, X. Yin,
M. Mazeika, A.-K. Dombrowski, S. Goel, N. Li, M. J. Byun, Z. Wang,
A. Mallen, S. Basart, S. Koyejo, D. Song, M. Fredrikson, J. Z. Kolter,
and D. Hendrycks,
``Representation engineering: A top-down approach to AI transparency,''
\textit{arXiv preprint arXiv:2310.01405}, 2023,
doi: \href{https://doi.org/10.48550/arXiv.2310.01405}{10.48550/arXiv.2310.01405}.
[Online]. Available: \url{https://arxiv.org/abs/2310.01405}.

\end{thebibliography}
\end{document}